\documentclass[10pt,twocolumn,letterpaper]{article}

\usepackage[pagenumbers]{cvpr}

\usepackage{microtype}

\usepackage{amsthm}
\newtheoremstyle{raceboxed}{0pt}{0pt}{\normalfont}{}{\bfseries}{.}{.5em}{\thmname{#1}\thmnumber{ #2}\thmnote{ (#3)}}
\newtheoremstyle{raceplain}{\topsep}{\topsep}{\normalfont}{}{\bfseries}{.}{.5em}{\thmname{#1}\thmnumber{ #2}\thmnote{ (#3)}}
\theoremstyle{raceboxed}
\newtheorem{proposition}{Proposition}
\theoremstyle{raceplain}
\newtheorem{lemma}{Lemma}

\newsavebox{\propositionbox}
\newenvironment{propositionframe}{%
  \par\addvspace{6pt}%
  \setlength{\fboxsep}{5.5pt}\setlength{\fboxrule}{0.45pt}%
  \begin{lrbox}{\propositionbox}%
  \begin{minipage}{\dimexpr\linewidth-2\fboxsep-2\fboxrule\relax}%
}{%
  \end{minipage}%
  \end{lrbox}%
  \noindent\fcolorbox{black!45}{white}{\usebox{\propositionbox}}%
  \par\addvspace{6pt}%
}
\BeforeBeginEnvironment{proposition}{\begin{propositionframe}}
\AfterEndEnvironment{proposition}{\end{propositionframe}}

\usepackage{multirow}
\usepackage{booktabs}
\usepackage[ruled,vlined]{algorithm2e}

\newcommand{\modelname}{RACE-AD\xspace}

\ifdefined\pdfsuppressptexinfo\pdfsuppressptexinfo=-1\fi

\definecolor{cvprblue}{rgb}{0.21,0.49,0.74}
\usepackage[pagebackref,breaklinks,colorlinks,allcolors=cvprblue]{hyperref}

\def\confName{CVPR}
\def\confYear{2027}

\title{Relation-Aware Evidence Routing and Confidence-Guided Adaptation \\for Zero-Shot Visual Anomaly Detection}

\author{
Xinyu Zhao \qquad Yang Hu \qquad Jiayi Luo \qquad Qingyun Sun \qquad Jianxin Li\\
School of Computer Science and Engineering, Beihang University\\
{\tt\small \{xyzhao,sunqy\}@buaa.edu.cn}
}

\begin{document}
\maketitle

\begin{abstract}
Zero-shot visual anomaly detection (ZSAD) aims to identify and localize anomalies in unseen categories without target-domain adaptation, yet heterogeneous anomaly patterns and cross-domain variations challenge reliable detection. Existing approaches primarily rely on local visual features, patch-text similarities, or memory-based retrieval, leaving inter-patch relational information underexplored in anomaly evidence modeling. In this paper, we propose RACE-AD, a relation-aware evidence routing and confidence-guided adaptation framework that integrates local anomaly responses with attention-derived relational information for generalizable ZSAD. To capture inter-patch relationships, we characterize the dispersion of visual attention distributions using Shannon entropy, providing complementary relational statistics beyond local anomaly evidence. To organize heterogeneous evidence, we introduce relation-aware routing that combines memory-derived normal/anomaly responses with attention entropy to construct image-conditioned representations through entropy-regularized soft aggregation. To mitigate unreliable retrieval responses under domain shift, we design confidence-guided adaptation that regulates anomaly-conditioned representations according to the strength of the memory anomaly evidence and the relational dispersion before visual-semantic alignment. Extensive experiments on ten industrial and medical benchmarks demonstrate the effectiveness of RACE-AD, achieving average image-level AUROC and AP scores of 91.9\% and 93.7\%, and pixel-level AUROC and AUPRO scores of 93.6\% and 80.0\%, respectively. Ablation and diagnostic analyses show that entropy-guided routing mainly improves region-level localization and that the confidence gate is essential under domain shift.
\end{abstract}

\section{Introduction}
\label{sec:introduction}

\begin{figure}[t]
    \centering
    \includegraphics[width=\linewidth]{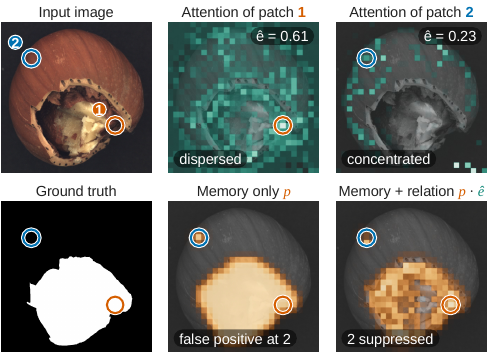}
    \caption{Patch~1 (crack) and patch~2 (normal surface) receive similar memory responses $p$ but distribute their attention differently. Memory evidence alone yields a false positive at patch~2; the joint evidence $p\cdot\hat{e}$ with attention entropy $\hat{e}$ retains the crack and suppresses it. Bottom: ground truth and illustrative evidence maps.}
    \label{fig:motivation}
\end{figure}

Visual anomaly detection identifies irregular patterns and localizes the corresponding regions, with applications in industrial inspection and medical imaging~\cite{bergmann2019mvtec,zou2022spot,fernando2021deep}. Conventional methods are trained for each category and require representative normal samples from the target domain~\cite{roth2022towards,deng2022anomaly,liu2023simplenet}, which is costly when anomalies are rare and target categories change frequently. Zero-shot visual anomaly detection (ZSAD) instead transfers anomaly knowledge learned from annotated source data to unseen categories without any target-domain adaptation. Its main difficulty is that anomaly appearance, spatial extent, and visual context vary substantially across domains.

Vision-language models, particularly CLIP~\cite{radford2021learning}, have made ZSAD practical through transferable visual-semantic representations. WinCLIP~\cite{jeong2023winclip} and AnomalyCLIP~\cite{zhouanomalyclip} separate normal and anomalous patterns with handcrafted or learnable prompts, while FiLo~\cite{gu2024filo} and FAPrompt~\cite{zhu2025fine} refine the alignment with fine-grained abnormality descriptions. AA-CLIP~\cite{ma2025aa} and MoECLIP~\cite{park2026moeclip} make visual representations more sensitive to anomalies, and MRAD~\cite{xu2026mrad} retrieves localized anomaly responses from a source-domain memory. In all of these methods, the anomaly evidence of a patch is computed from the patch itself, namely from its feature, its similarity to text, or its match in memory. How the patch interacts with the rest of the image is not used explicitly.

Fig.~\ref{fig:motivation} illustrates why this relational information is useful. Memory retrieval measures how well a patch matches learned normal or anomalous patterns, whereas self-attention records which other regions the patch interacts with. Two patches can receive similar memory responses and still distribute their attention very differently: the crack patch in Fig.~\ref{fig:motivation} attends to the whole image, while the normal patch attends to a few locations. Attention entropy summarizes this difference with a single number per patch. Its magnitude, however, is not an anomaly score, since many normal regions also have high entropy; using it as independent evidence confuses relational variation with abnormality. A second difficulty concerns the reliability of the memory. On unseen categories, unfamiliar but normal patterns can receive high anomaly responses, as patch~2 does in Fig.~\ref{fig:motivation}, and such responses propagate into image-conditioned prompt learning and distort the resulting visual-semantic alignment. These observations call for organizing local and relational evidence jointly and for controlling how strongly uncertain anomaly evidence influences the alignment.

We propose \textbf{RACE-AD}, a relation-aware evidence routing and confidence-guided adaptation framework for ZSAD. RACE-AD computes the Shannon entropy of each patch's attention distribution in the frozen CLIP encoder and uses it alongside memory-derived anomaly evidence. Relation-aware evidence routing weights the two signals jointly and pools patch features into an image-conditioned normal token and an anomaly token. The routing weights are the unique maximizer of an entropy-regularized evidence selection objective, so a single temperature trades evidence preference against aggregation diversity. Confidence-guided adaptation then scales the anomaly token by a gate computed from the strongest memory response and the spatial variation of attention entropy, before the two tokens condition learnable text prompts. We further prove that the displacement of the routed anomaly token caused by memory errors is bounded in proportion to the attention entropy of the affected patches, and that the gate bounds how far weak anomaly evidence can shift the anomaly prompt. On ten industrial and medical benchmarks, RACE-AD achieves average image-level AUROC/AP of 91.9\%/93.7\% and pixel-level AUROC/AUPRO of 93.6\%/80.0\%.

Our main contributions are as follows:
\begin{itemize}
    \item We introduce a relation-aware view of ZSAD that complements memory-based anomaly evidence with attention-derived relational statistics, so that representations reflect both local responses and inter-patch interactions.
    \item We propose RACE-AD, which pools normal and anomaly tokens through entropy-regularized evidence routing and regulates the anomaly token with a confidence gate before visual-semantic alignment.
    \item On ten industrial and medical benchmarks, RACE-AD obtains the best average detection and localization results among the compared methods, and ablation and diagnostic analyses characterize the contribution of each component.
\end{itemize}

\section{Related Work}
\label{sec:related}

\subsection{Zero-Shot Visual Anomaly Detection}
\label{sec:related_zsad}

Unsupervised methods require normal images of each target category~\cite{deng2022anomaly,liu2023simplenet,zhang2024realnet,guo2025dinomaly,wang2025cnc,defard2021padim,roth2022towards,hyun2024reconpatch}, whereas ZSAD methods build on the transferability of CLIP~\cite{radford2021learning} to dense prediction~\cite{zhong2022regionclip,yao2023detclipv2,wang2022cris,liu2023clip}. WinCLIP~\cite{jeong2023winclip} combines handcrafted prompts with multi-scale windows, AnomalyCLIP~\cite{zhouanomalyclip} learns object-agnostic prompts, AdaCLIP~\cite{cao2024adaclip} adds image-conditioned prompts, and FiLo~\cite{gu2024filo} and FAPrompt~\cite{zhu2025fine} use fine-grained abnormality descriptions. APRIL-GAN~\cite{chen2023april}, AA-CLIP~\cite{ma2025aa}, AF-CLIP~\cite{fang2025af}, and AdaptCLIP~\cite{gao2025adaptclip} adapt the visual or cross-modal representations, PromptMoE~\cite{shao2025promptmoe} and MoECLIP~\cite{park2026moeclip} assign specialized prompts or experts with a learned router~\cite{jacobs1991adaptive,shazeer2017outrageously}, and MRAD~\cite{xu2026mrad} retrieves anomaly evidence from memories built on annotated source data. These methods compute the anomaly evidence of each patch independently, so an unfamiliar but normal patch can respond as strongly as a defect, and prompts conditioned on such responses inherit the error. RACE-AD adds relational evidence to the image-conditioned prompts, and its routing pools patches into two evidence tokens with closed-form weights instead of a learned router.

\subsection{Entropy of Relational Structure}
\label{sec:related_attention}

Self-attention in vision transformers~\cite{dosovitskiy2020image,vaswani2017attention} records how each patch interacts with the other patches. The Shannon entropy~\cite{shannon1948mathematical} of attention distributions has been used to analyze and stabilize transformer training, where a collapse of attention entropy accompanies training instability~\cite{zhai2023stabilizing}, and large vision transformers repurpose a few background tokens for global computation, which then receive much of the attention~\cite{darcet2024vision}. On graphs, structural entropy~\cite{li2016structural,su2025survey} has guided pooling, contrastive learning, and structure learning~\cite{wu2022structural,wu2023sega,zou2023se}. A row-normalized attention matrix is the transition matrix of a random walk over patches, and we use the entropy of each row as a relational statistic, only in combination with memory evidence.

\section{Methodology}
\label{sec:method}

\subsection{Problem Formulation}
\label{sec:problem}

Let $\mathcal{X} \subset \mathbb{R}^{H \times W \times C}$ denote the image space. In ZSAD, a model $F_{\Theta}$ is trained on a source-domain dataset $\mathcal{D}_{\mathrm{train}} = \{(x_i, y_i, \mathbf{Y}_i)\}_{i=1}^{N_{\mathrm{train}}}$, where $x_i \in \mathcal{X}$ is an image, $y_i \in \{0,1\}$ is its image-level label, and $\mathbf{Y}_i \in \{0,1\}^{H \times W}$ is its pixel-level mask. The model is then evaluated directly on target datasets $\{\mathcal{D}_{\mathrm{test}}^1, \dots, \mathcal{D}_{\mathrm{test}}^{n}\}$ that are not used for training, without any target-domain adaptation. For a test image $x \in \mathcal{D}_{\mathrm{test}}^j$, the model predicts an image-level anomaly score $a \in \mathbb{R}$ and a pixel-level localization map $\mathbf{M} \in \mathbb{R}^{H \times W}$, denoted as $(a, \mathbf{M}) = F_{\Theta}(x)$.

\begin{figure*}[t]
    \centering
    \includegraphics[width=\textwidth]{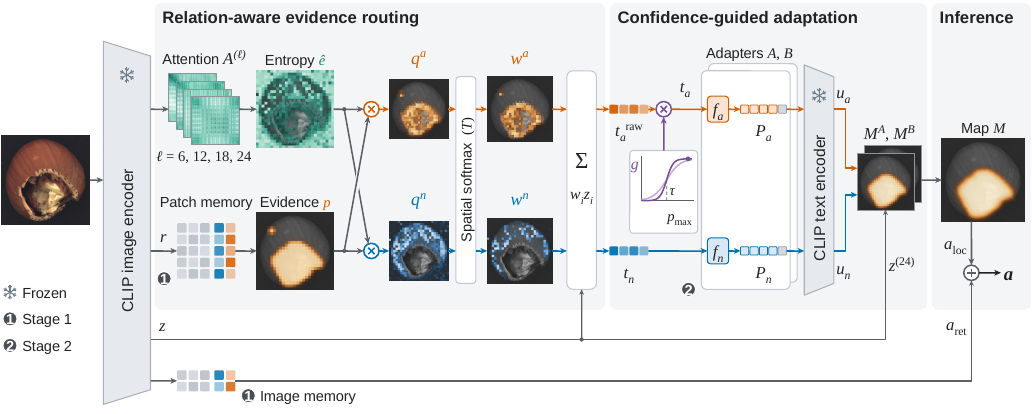}
    \caption{Overview of RACE-AD. Relation-aware evidence routing combines the attention entropy $\hat{e}$ of the frozen CLIP encoder with the memory evidence $p$ to pool patch features into the tokens $t_n$ and $t_a^{\mathrm{raw}}$. Confidence-guided adaptation scales the anomaly token by the gate $g$ before two prompt adapters condition the learnable prompts. The resulting anomaly maps are fused into $M$, whose top response is combined with the retrieval score $a_{\mathrm{ret}}$.}
    \label{fig:framework}
\end{figure*}

\subsection{Framework Overview}
\label{sec:overview}

Fig.~\ref{fig:framework} gives an overview of RACE-AD. For each image, a frozen CLIP image encoder provides the self-attention matrices of the layers in $\Omega$, projected patch features for memory retrieval, and dense patch features. Training has two stages. Stage~1 learns image-level and patch-level memory retrieval, which yields an image-level retrieval score and patch-wise normal and anomaly responses. Stage~2 keeps the encoder and the Stage-1 modules frozen and learns the prompt-related modules: relation-aware evidence routing (\cref{sec:routing}) combines memory responses with attention entropy to pool patch features into a normal token and an anomaly token, and confidence-guided adaptation (\cref{sec:adaptation}) regulates the anomaly token before the two tokens condition learnable text prompts. At inference (\cref{sec:training_inference}), the prompt-based anomaly maps form the localization map, whose strongest responses are fused with the image-level retrieval score. Along the way, we characterize how the utilities, the routing, and the gate respond to unreliable memory evidence; all proofs are given in Appendix~\ref{sec:supp_proof}.

\subsection{Relation-Aware Evidence Routing}
\label{sec:routing}

Local memory responses characterize visual deviations at individual patches, while attention distributions describe how patches interact. We combine the two signals to construct image-conditioned normal and anomaly representations.

\noindent\textbf{Attention-Based Relational Statistics.}
Let $\mathbf{A}^{(\ell)}\in\mathbb{R}^{(N+1)\times(N+1)}$ denote the head-averaged self-attention matrix at visual layer $\ell$, where $N$ is the number of spatial patches and the additional token is the global token. After removing the global token, we renormalize each spatial attention row:
\begin{equation}
\widetilde{A}_{ij}^{(\ell)}
=
\frac{A_{ij}^{(\ell)}}
{\sum_{k=1}^{N}A_{ik}^{(\ell)}+\epsilon}.
\label{eq:attention}
\end{equation}
The normalized row is the distribution of attention that patch $i$ assigns to the spatial locations. We measure its dispersion with the Shannon entropy
\begin{equation}
e_i^{(\ell)}
=
-\sum_{j=1}^{N}
\widetilde{A}_{ij}^{(\ell)}
\log\left(\widetilde{A}_{ij}^{(\ell)}+\epsilon\right).
\label{eq:attention_entropy}
\end{equation}
Low entropy means that the patch attends to a few locations, and high entropy that it interacts with many regions. Averaging over the layers $\Omega=\{6,12,18,24\}$, $e_i=|\Omega|^{-1}\sum_{\ell\in\Omega}e_i^{(\ell)}$, gives a per-patch statistic of relational context that requires no additional parameters.

\noindent\textbf{Memory-Derived Anomaly Evidence.}
The source-domain patch memory stores $M$ prototype features with one-hot normal and anomaly values $\mathbf{V}_{\mathrm{pat}}\in\mathbb{R}^{M\times 2}$. Two learned linear projections map the feature of patch $i$ to $\mathbf{r}_i\in\mathbb{R}^{d}$ and the prototypes to the keys $\mathbf{K}_{\mathrm{pat}}\in\mathbb{R}^{M\times d}$. With the similarity vector $\mathbf{s}_i=\mathbf{K}_{\mathrm{pat}}\mathbf{r}_i$, the memory response is
\begin{equation}
\mathbf{o}_i
=
\operatorname{softmax}(\mathbf{s}_i)^\top
\mathbf{V}_{\mathrm{pat}},
\label{eq:memory_response}
\end{equation}
where $\mathbf{o}_i=[o_{i,n},o_{i,a}]$ contains the normal and anomaly responses, and the patch-wise anomaly evidence is $p_i=o_{i,a}$. Quantile-based similarity filtering is applied during source-domain training and disabled at inference.

\noindent\textbf{Joint Evidence Utilities.}
We normalize the attention entropy within each image, so that $\hat e_i$ is comparable across categories,
\begin{equation}
\hat e_i
=
\frac{e_i-\min_j e_j}
{\max_j e_j-\min_j e_j+\epsilon},
\label{eq:entropy_normalization}
\end{equation}
and define the anomaly and normal routing utilities as $q_i^a=p_i\hat e_i$ and $q_i^n=(1-p_i)(1-\hat e_i)$. The two utilities have a simple structure.

\begin{proposition}[Evidence agreement]\leavevmode\label{prop:agreement}%
For every patch $i$, $q_i^a\le\min(p_i,\hat e_i)$, $q_i^n\le\min(1-p_i,1-\hat e_i)$, and
\begin{equation}
q_i^a-q_i^n=p_i+\hat e_i-1.
\label{eq:agreement}
\end{equation}
\end{proposition}

A patch thus receives a large utility only when memory evidence and relational statistics agree, and its anomaly utility exceeds its normal utility exactly when $p_i+\hat e_i>1$. Entropy alone cannot make a patch anomalous: since $q_i^a\le p_i$, a patch that the memory considers normal keeps a small anomaly utility for any $\hat e_i$. An additive utility $(p_i+\hat e_i)/2$ would instead give such a patch up to half of the maximal value. Among patches with strong anomaly responses, anomaly routing prefers those with dispersed attention, which attenuates false responses on patches with concentrated attention, such as patch~2 in Fig.~\ref{fig:motivation}; normal routing prefers memory-supported normal patches with concentrated attention.

\noindent\textbf{Entropy-Regularized Routing.}
For each evidence type $c\in\{n,a\}$, we select patches by solving
\begin{equation}
\mathbf{w}^{c*}
=
\underset{\mathbf{w}\in\Delta_N}{\arg\max}\;
J(\mathbf{w}),
\quad
J(\mathbf{w})
=
\sum_{i=1}^{N}w_iq_i^c
+
T\mathcal{H}(\mathbf{w}),
\label{eq:routing_objective}
\end{equation}
where $\Delta_N$ is the probability simplex, $\mathcal{H}(\mathbf{w})=-\sum_iw_i\log w_i$ is the entropy of the routing distribution, and $T>0$ is the routing temperature. The first term favors patches with stronger joint evidence, and the second keeps the aggregation from collapsing onto a single patch. The solution is available in closed form.

\begin{proposition}[Optimal evidence routing]\leavevmode\label{prop:routing}%
For $T>0$, $J$ is strictly concave over $\Delta_N$, and its unique maximizer is the spatial softmax
\begin{equation}
w_i^{c*}
=
\frac{\exp(q_i^c/T)}
{\sum_{j=1}^{N}\exp(q_j^c/T)}.
\label{eq:routing_weights}
\end{equation}
Moreover, for any $\mathbf{w}\in\Delta_N$,
\begin{equation}
J(\mathbf{w}^{c*})-J(\mathbf{w})
=
T D_{\mathrm{KL}}(\mathbf{w}\,\|\,\mathbf{w}^{c*})
\geq 0.
\label{eq:routing_optimality}
\end{equation}
\end{proposition}

Proposition~\ref{prop:routing} is the variational characterization of the softmax~\cite{boyd2004convex}. We use $\mathbf{w}^c=\mathbf{w}^{c*}$ as routing weights. The temperature interpolates between selecting the single patch with the largest utility ($T\to0$) and uniform pooling ($T\to\infty$), and Eq.~\eqref{eq:routing_optimality} shows that any other weighting loses objective value in proportion to its divergence from $\mathbf{w}^{c*}$. Given the $\ell_2$-normalized dense visual features $\{\mathbf{z}_i\}_{i=1}^{N}$, we construct the normal and raw anomaly tokens
\begin{equation}
\mathbf{t}_n
=
\sum_{i=1}^{N}w_i^n\mathbf{z}_i,
\qquad
\mathbf{t}_a^{\mathrm{raw}}
=
\sum_{i=1}^{N}w_i^a\mathbf{z}_i,
\label{eq:routed_tokens}
\end{equation}
which summarize, for each image, the patches selected by the joint evidence.

\noindent\textbf{Stability to Memory Errors.}
Memory responses on unseen categories are noisy, so we ask how much an error in $\mathbf{p}$ can move the routed tokens. Let $R$ be the radius of the smallest ball that contains $\{\mathbf{z}_i\}_{i=1}^{N}$.

\begin{proposition}[Stability of routed tokens]\leavevmode\label{prop:stability}%
Let the memory evidence be perturbed to $\mathbf{p}'=\mathbf{p}+\boldsymbol{\delta}\in[0,1]^N$ while $\hat{\mathbf{e}}$ and $\{\mathbf{z}_i\}$ are fixed, and let primes denote the resulting routing weights and tokens. Then
\begin{equation}
\|\mathbf{t}_a'^{\,\mathrm{raw}}-\mathbf{t}_a^{\mathrm{raw}}\|_2
\le R\,\|\mathbf{w}'^{a}-\mathbf{w}^{a}\|_1
\le \frac{R}{T}\max_i\hat e_i|\delta_i|,
\label{eq:stability}
\end{equation}
and the same bound holds for $\mathbf{t}_n$ with $\hat e_i$ replaced by $1-\hat e_i$.
\end{proposition}

A memory error at a patch with concentrated attention therefore barely moves the anomaly token, as for patch~2 in Fig.~\ref{fig:motivation}; it mainly removes the patch from the normal token, whose sensitivity is governed by $1-\hat e_i$. The factor $1/T$ quantifies the cost of selective routing: a smaller temperature concentrates the weights but amplifies memory errors, which matches the degradation observed for small $T$ in \cref{sec:analysis}.

Apart from the Stage-1 memory, the routing introduces no learnable parameters. The attention matrices are by-products of the frozen encoder, and the entropies and routing weights take $\mathcal{O}(|\Omega|N^2+Nd)$ operations per image, which is small compared with the encoder's own attention layers.

\subsection{Confidence-Guided Adaptation}
\label{sec:adaptation}

The reliability of memory-derived anomaly evidence varies across unseen categories. Confidence-guided adaptation regulates how strongly the routed anomaly evidence influences visual-semantic alignment.

\noindent\textbf{Confidence-Guided Token Regulation.}
We measure the strength of the anomaly evidence by $p_{\max}=\max_i p_i$ and the relational dispersion by $\sigma_e=\operatorname{Std}(\hat{\mathbf{e}})$. We use the maximum because a defect often covers few of the $N$ patches, which an average or a quantile of $\mathbf{p}$ would dilute. The confidence gate is
\begin{equation}
g
=
\sigma\left(
(p_{\max}-\tau)(k_0+k_1\sigma_e)
\right),
\label{eq:confidence_gate}
\end{equation}
where $\sigma(\cdot)$ is the sigmoid function, $\tau$ is a confidence threshold, and $k_0,k_1>0$ are fixed scalars. The regulated anomaly token is $\mathbf{t}_a=g\,\mathbf{t}_a^{\mathrm{raw}}$, while the normal token $\mathbf{t}_n$ is unchanged. The gate scales the anomaly token as a whole and the routing selects its patches, so a normal patch with a high memory response but concentrated attention contributes little to $\mathbf{t}_a$ even when the gate is open.

\noindent\textbf{Image-Conditioned Prompt Adaptation.}
The normal and regulated anomaly tokens condition learnable text prompts. We use two independently parameterized prompt adapters $m\in\{A,B\}$ that receive the same token pair. Each adapter generates prompt biases $\mathbf{b}_n^m=f_n^m(\mathbf{t}_n)$ and $\mathbf{b}_a^m=f_a^m(\mathbf{t}_a)$ with linear projections $f_c^m(\mathbf{t})=\mathbf{W}_c^m\mathbf{t}+\mathbf{c}_c^m$, and adds them to the learnable context embeddings of the normal and anomaly prompts. The frozen CLIP text encoder then produces normalized text representations
\begin{equation}
\mathbf{u}_c^m
=
\operatorname{Norm}
\left(
\operatorname{Enc}_{t}
\left(\mathcal{P}_c(\mathbf{b}_c^m)\right)
\right),
\quad c\in\{n,a\},
\label{eq:adapted_prompts}
\end{equation}
where $\mathcal{P}_c(\mathbf{b}_c^m)$ denotes the prompt of class $c$ whose context embeddings are shifted by $\mathbf{b}_c^m$. The gate controls how far the anomaly prompt can move away from its image-independent version $\mathcal{P}_a(\mathbf{c}_a^m)$. Let $B=\max_i\|\mathbf{z}_i\|_2$, which equals 1 for the normalized features.

\begin{proposition}[Gate attenuation]\leavevmode\label{prop:gate}%
The gate $g$ increases with $p_{\max}$, and $\partial g/\partial\sigma_e$ has the sign of $p_{\max}-\tau$. If $p_{\max}\le\tau-\Delta$ for some $\Delta>0$, then $g\le\sigma\left(-\Delta(k_0+k_1\sigma_e)\right)$ and
\begin{equation}
\|\mathbf{b}_a^m-\mathbf{c}_a^m\|_2
\le\|\mathbf{W}_a^m\|_2\,B\,\sigma\left(-\Delta(k_0+k_1\sigma_e)\right).
\label{eq:gate_bound}
\end{equation}
\end{proposition}

When no patch has a confident anomaly response, the anomaly prompt stays close to its image-independent version, so unreliable retrieval has little influence on the visual-semantic alignment. With $\tau=0.5$, $k_0=5$, and $k_1=50$, a margin $\Delta=0.2$ already gives $g\le0.27$, and $g\le0.05$ once $\sigma_e\ge0.2$. Larger relational dispersion sharpens the gate in both directions, pushing $g$ toward 1 for confident images and toward 0 otherwise.

\noindent\textbf{Patch-Text Alignment.}
For localization, we use the dense patch features of the final visual layer $\ell^*=24$. With $s_{i,c}^m=\operatorname{sim}(\mathbf{z}_i^{(24)},\mathbf{u}_c^m)$ the cosine similarity between patch $i$ and text representation $c$, the patch-level anomaly response is the two-class softmax
\begin{equation}
S_{a,i}^{m}
=
\frac{\exp(s_{i,a}^m/0.07)}
{\exp(s_{i,n}^m/0.07)+\exp(s_{i,a}^m/0.07)},
\label{eq:patch_text_alignment}
\end{equation}
with a fixed temperature of 0.07, and the normal response is $S_{n,i}^{m}=1-S_{a,i}^{m}$. Reshaping the patch-wise responses to the spatial grid and resizing them to the input resolution gives the anomaly map $\mathbf{M}^m$; both responses are used for source-domain supervision.


\begin{table*}[t]
\centering
\begingroup
\footnotesize
\setlength{\tabcolsep}{0.4mm}
\renewcommand{\arraystretch}{1.05}
\begin{tabular*}{\textwidth}{@{\extracolsep{\fill}}lccccccccc@{}}
\toprule
\multirow{2}{*}{Method} & \multicolumn{5}{c}{Industrial} & \multicolumn{3}{c}{Medical} & \multirow{2}{*}{Avg.} \\
\cmidrule(lr){2-6}\cmidrule(lr){7-9}
& MVTec-AD & VisA & BTAD & DTD & MPDD & BrainMRI & HeadCT & Br35H & \\
\midrule
WinCLIP~\cite{jeong2023winclip}
& 90.4/95.6 & 75.5/78.7 & 68.2/70.9 & 95.1/97.7 & 61.5/69.2 & 86.5/91.5 & 81.8/80.2 & 81.0/83.2 & 80.0/83.4 \\

AnomalyCLIP~\cite{zhouanomalyclip}
& 88.1/95.1 & 78.4/82.2 & 81.8/77.0 & 93.5/97.0 & 70.9/79.1 & 88.5/91.3 & 87.8/86.9 & 89.5/90.0 & 84.8/87.3 \\

FiLo~\cite{gu2024filo}
& 87.4/94.3 & 79.6/84.2 & 84.1/92.0 & 94.1/98.0 & 72.2/74.9 & 86.0/85.6 & 93.8/93.4 & 93.6/92.8 & 86.4/89.4 \\

AA-CLIP~\cite{ma2025aa}
& 89.7/94.8 & 78.5/82.0 & \textbf{93.3}/\underline{96.6} & 94.3/93.7 & 57.7/67.0 & 92.6/94.9 & 88.4/90.9 & 89.6/91.1 & 85.5/88.9 \\

FAPrompt~\cite{zhu2025fine}
& 91.0/95.2 & 82.5/84.3 & 91.0/89.7 & \textbf{96.0}/\underline{98.3} & 77.9/\underline{81.7} & \underline{95.7}/95.7 & 93.4/92.9 & \underline{96.8}/\underline{96.6} & 90.5/91.8 \\

MoECLIP~\cite{park2026moeclip}
& 91.4/96.1 & 80.9/84.5 & 90.8/\textbf{97.1} & 93.3/97.7 & 68.6/69.9 & 91.9/94.1 & 92.5/93.4 & 91.6/90.9 & 87.6/90.5 \\

MRAD~\cite{xu2026mrad}
& \underline{93.0}/\underline{97.0} & \underline{83.2}/\underline{85.5} & 92.7/90.5 & 95.0/98.1 & \underline{78.4}/80.3 & 95.3/\underline{96.2} & \underline{94.8}/\underline{95.4} & 96.3/96.5 & \underline{91.1}/\underline{92.4} \\

\midrule
\textbf{RACE-AD (Ours)}
& \textbf{93.4}/\textbf{97.3} & \textbf{84.7}/\textbf{87.2} & \underline{92.8}/94.0 & \underline{95.9}/\textbf{98.5} & \textbf{79.3}/\textbf{82.6} & \textbf{96.1}/\textbf{96.9} & \textbf{95.8}/\textbf{96.5} & \textbf{96.9}/\textbf{96.9} & \textbf{91.9}/\textbf{93.7} \\
\bottomrule
\end{tabular*}
\endgroup
\caption{Cross-dataset zero-shot anomaly detection, AUROC/AP (\%). Best and second-best results are in \textbf{bold} and \underline{underlined}.}
\label{tab:zero-shot-image}
\end{table*}

\begin{table*}[t]
\centering
\begingroup
\footnotesize
\setlength{\tabcolsep}{0.4mm}
\renewcommand{\arraystretch}{1.05}
\begin{tabular*}{\textwidth}{@{\extracolsep{\fill}}lcccccccc@{}}
\toprule
\multirow{2}{*}{Method} & \multicolumn{5}{c}{Industrial} & \multicolumn{2}{c}{Medical} & \multirow{2}{*}{Avg.} \\
\cmidrule(lr){2-6}\cmidrule(lr){7-8}
& MVTec-AD & VisA & BTAD & DTD & MPDD & Endo & Kvasir & \\
\midrule
WinCLIP~\cite{jeong2023winclip}
& 82.3/61.9 & 73.2/51.0 & 72.7/27.3 & 79.5/51.5 & 71.2/40.5 & 68.2/28.3 & 69.7/24.5 & 73.8/40.7 \\

AnomalyCLIP~\cite{zhouanomalyclip}
& 89.5/83.0 & 94.4/86.7 & 90.8/66.4 & 97.9/92.3 & 96.3/90.0 & 83.4/61.2 & 79.3/52.0 & 90.2/75.9 \\

FiLo~\cite{gu2024filo}
& 90.2/57.2 & \textbf{95.7}/80.6 & 90.3/52.8 & 98.0/77.5 & 93.9/69.1 & 87.5/66.2 & \underline{85.0}/51.7 & 91.5/65.0 \\

AA-CLIP~\cite{ma2025aa}
& 91.9/84.6 & 94.8/81.4 & 93.9/69.8 & 97.0/88.7 & 95.9/86.4 & 88.0/68.7 & 82.7/48.9 & 92.0/75.5 \\

FAPrompt~\cite{zhu2025fine}
& 90.7/82.1 & \textbf{95.7}/\underline{86.8} & 95.9/\underline{75.3} & 98.2/91.8 & 95.6/85.4 & 85.5/65.3 & 80.8/46.7 & 91.8/76.2 \\

MoECLIP~\cite{park2026moeclip}
& 92.4/\textbf{87.0} & \underline{95.0}/85.7 & 93.6/68.8 & \underline{98.4}/\textbf{92.8} & 96.4/87.7 & 88.3/69.0 & \textbf{85.3}/51.4 & 92.8/77.5 \\

MRAD~\cite{xu2026mrad}
& \underline{92.6}/84.9 & 94.9/84.9 & \textbf{96.9}/\textbf{75.9} & 98.1/89.8 & \underline{97.4}/\underline{90.4} & \underline{89.1}/\underline{73.7} & 84.9/\underline{52.7} & \underline{93.4}/\underline{78.9} \\

\midrule
\textbf{RACE-AD (Ours)}
& \textbf{92.8}/\underline{86.4} & 94.8/\textbf{87.2} & \underline{96.2}/74.3 & \textbf{98.5}/\underline{92.5} & \textbf{97.9}/\textbf{92.3} & \textbf{89.4}/\textbf{73.9} & \textbf{85.3}/\textbf{53.5} & \textbf{93.6}/\textbf{80.0} \\
\bottomrule
\end{tabular*}
\endgroup
\caption{Cross-dataset zero-shot anomaly localization, AUROC/AUPRO (\%). Best and second-best results are in \textbf{bold} and \underline{underlined}.}
\label{tab:zero-shot-pixel}
\end{table*}

\subsection{Training and Inference}
\label{sec:training_inference}

\noindent\textbf{Two-Stage Training.}
RACE-AD is trained only on labeled source-domain data with a frozen CLIP visual backbone. In Stage~1, the image-level and patch-level memory retrieval modules are optimized with image-level cross-entropy and pixel-level segmentation supervision:
\begin{equation}
\mathcal{L}_{\mathrm{stage1}}
=
\mathcal{L}_{\mathrm{CE}}(\mathbf{h}_{\mathrm{img}}/0.07,y)
+
\mathcal{L}_{\mathrm{seg}}
(\widehat{\mathbf{M}}_{\mathrm{base}},\mathbf{Y}),
\label{eq:stage1_loss}
\end{equation}
where $\mathbf{h}_{\mathrm{img}}$ denotes the image-level logits before temperature scaling and $\widehat{\mathbf{M}}_{\mathrm{base}}$ the patch-level retrieval map resized to the input resolution. The segmentation objective combines a two-class focal loss with Dice supervision on the anomaly channel, $\mathcal{L}_{\mathrm{seg}}=\mathcal{L}_{\mathrm{focal}}^{(2)}+\lambda_d\mathcal{L}_{\mathrm{dice}}$.

In Stage~2, the visual backbone and the Stage-1 projection modules remain frozen, and the prompt learner and the adaptation modules are optimized. For each adapter $m\in\{A,B\}$, the objective combines two-class focal supervision with Dice losses on the anomaly and normal channels:
\begin{equation}
\begin{aligned}
\mathcal{L}_{\mathrm{clip}}^m
={}&\mathcal{L}_{\mathrm{focal}}^{(2)}
(\widehat{\mathbf{S}}^m,\mathbf{Y})
+\lambda_d\mathcal{L}_{\mathrm{dice}}
(\widehat{\mathbf{M}}_a^m,\mathbf{Y})\\
&+\lambda_d\mathcal{L}_{\mathrm{dice}}
(\widehat{\mathbf{M}}_n^m,1-\mathbf{Y}),
\end{aligned}
\label{eq:branch_loss}
\end{equation}
where $\widehat{\mathbf{S}}^m=[\widehat{\mathbf{M}}_n^m,\widehat{\mathbf{M}}_a^m]$ stacks the normal and anomaly maps at the input resolution. The Stage-2 objective is
\begin{equation}
\mathcal{L}_{\mathrm{stage2}}
=
\bar{\lambda}_A\mathcal{L}_{\mathrm{clip}}^A
+
\bar{\lambda}_B\mathcal{L}_{\mathrm{clip}}^B,
\label{eq:stage2_loss}
\end{equation}
with loss weights $\bar{\lambda}_m=\lambda_m/(\lambda_A+\lambda_B+\epsilon)$.

\noindent\textbf{Zero-Shot Inference.}
RACE-AD is applied to the target datasets without any target-domain adaptation. The anomaly maps of the two adapters are averaged and smoothed with a Gaussian filter of standard deviation $\sigma=4$ to obtain the localization map $\mathbf{M}$. For image-level detection, we average the top-$1\%$ responses of $\mathbf{M}$,
\begin{equation}
a_{\mathrm{loc}}
=
\frac{1}{|\mathcal{T}|}
\sum_{(u,v)\in\mathcal{T}}M_{uv},
\label{eq:localization_score}
\end{equation}
where $\mathcal{T}$ contains the spatial locations with the highest anomaly responses, and combine the result with the image-level retrieval score $a_{\mathrm{ret}}$:
\begin{equation}
a
=
(1-\kappa)a_{\mathrm{loc}}
+
\kappa a_{\mathrm{ret}},
\label{eq:final_score}
\end{equation}
with $\kappa=0.7$. All settings are fixed across target datasets.

\section{Experiments}
\label{sec:experiments}

\begin{figure*}[t]
    \centering
    \includegraphics[width=\textwidth]{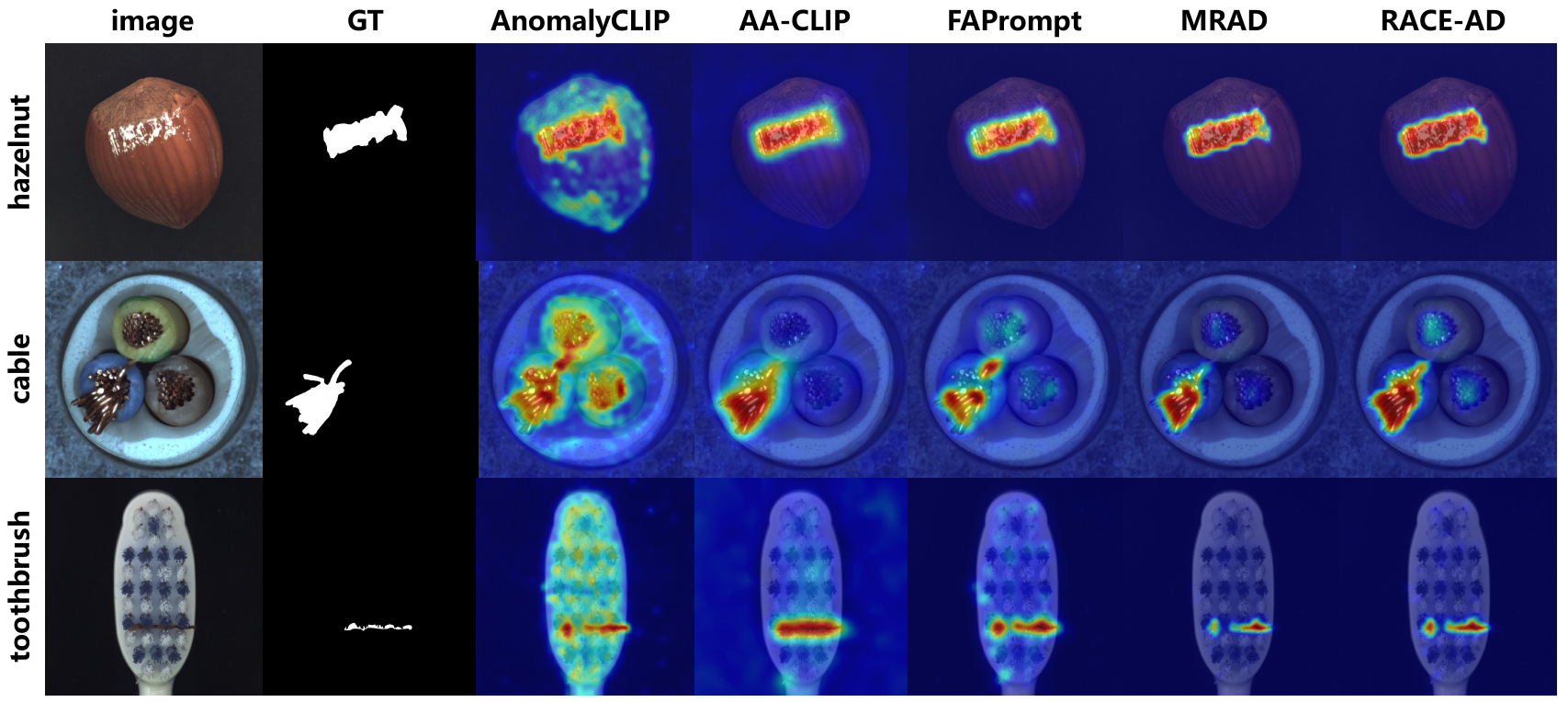}
    \caption{Qualitative comparison of zero-shot anomaly localization on MVTec-AD.}
    \label{fig:qualitative_comparison}
\end{figure*}

\subsection{Experimental Setup}
\label{sec:setup}

\noindent\textbf{Datasets and Protocol.}
We evaluate RACE-AD on ten industrial and medical anomaly detection benchmarks. The industrial datasets are MVTec-AD~\cite{bergmann2019mvtec}, VisA~\cite{zou2022spot}, BTAD~\cite{mishra2021vt}, DTD-Synthetic (DTD)~\cite{aota2023zero}, and MPDD~\cite{jezek2021deep}. The medical datasets are BrainMRI~\cite{kanade2015brain}, HeadCT~\cite{salehi2021multiresolution}, Br35H~\cite{hamada2020br35h}, Endo~\cite{hicks2021endotect}, and Kvasir~\cite{jha2019kvasir}. Following the cross-dataset ZSAD protocol~\cite{zhu2025fine}, we train on MVTec-AD and evaluate directly on the other nine datasets, and evaluate MVTec-AD with a model trained on VisA. Source-domain training uses the annotated test split of the source dataset, which contains normal and anomalous images with pixel-level masks. Target datasets are not used for training or adaptation. MVTec-AD and VisA share the capsule category; excluding it changes the RACE-AD averages on both by at most 0.8 points.

\noindent\textbf{Evaluation Metrics.}
Image-level detection is evaluated with AUROC and average precision (AP), and pixel-level localization with AUROC and the area under the per-region overlap curve (AUPRO). We report macro averages over datasets to summarize performance across heterogeneous target domains.

\noindent\textbf{Implementation Details.}
We use a frozen CLIP ViT-L/14@336px visual encoder with an input resolution of $518\times518$. Training takes one epoch for memory-guided retrieval and five epochs for image-conditioned prompt adaptation, with Adam, a learning rate of $4\times10^{-4}$, and a batch size of 8. The routing temperature is $T=0.1$, and the confidence gate uses $\tau=0.5$, $k_0=5.0$, and $k_1=50.0$. The two prompt adapters are trained with loss weights $(\lambda_A,\lambda_B)=(0.7,0.3)$ and averaged with equal weights at inference. The same configuration is used for every target dataset; Appendix~\ref{sec:supp_implementation} gives further implementation details, and \cref{sec:analysis} studies the sensitivity to $T$ and $(\lambda_A,\lambda_B)$.

\subsection{Comparison with State-of-the-Art Methods}
\label{sec:sota}

We compare RACE-AD with representative CLIP-based ZSAD methods: WinCLIP~\cite{jeong2023winclip}, AnomalyCLIP~\cite{zhouanomalyclip}, FiLo~\cite{gu2024filo}, AA-CLIP~\cite{ma2025aa}, FAPrompt~\cite{zhu2025fine}, MoECLIP~\cite{park2026moeclip}, and MRAD~\cite{xu2026mrad}. All baseline results follow the same cross-dataset protocol. Tables~\ref{tab:zero-shot-image} and~\ref{tab:zero-shot-pixel} report the image-level and pixel-level results.

\noindent\textbf{Image-Level Anomaly Detection.}
As shown in Table~\ref{tab:zero-shot-image}, RACE-AD obtains the best average AUROC and AP, 91.9\% and 93.7\%, which are 0.8 and 1.3 points higher than those of MRAD, the strongest baseline. It ranks first in AUROC on six of the eight datasets and in AP on seven, including all three medical datasets, and its AUROC on BTAD and DTD is within 0.5 points of the best result. The gains hold for both transfer directions, since MVTec-AD is evaluated with the VisA-trained model and the other datasets with the MVTec-AD-trained model.

\noindent\textbf{Pixel-Level Anomaly Localization.}
Table~\ref{tab:zero-shot-pixel} shows that RACE-AD achieves average AUROC and AUPRO of 93.6\% and 80.0\%, improving on MRAD by 0.2 and 1.1 points. The improvement is larger for AUPRO, which weights every anomalous region equally regardless of its size: RACE-AD has the best AUPRO on VisA, MPDD, Endo, and Kvasir and the best AUROC on MVTec-AD, DTD, MPDD, Endo, and Kvasir. MRAD is stronger on BTAD, and MoECLIP obtains a higher AUPRO on MVTec-AD and DTD.

\subsection{Ablation Studies}
\label{sec:ablation}

\begin{figure}[t]
    \centering
    \includegraphics[width=\linewidth]{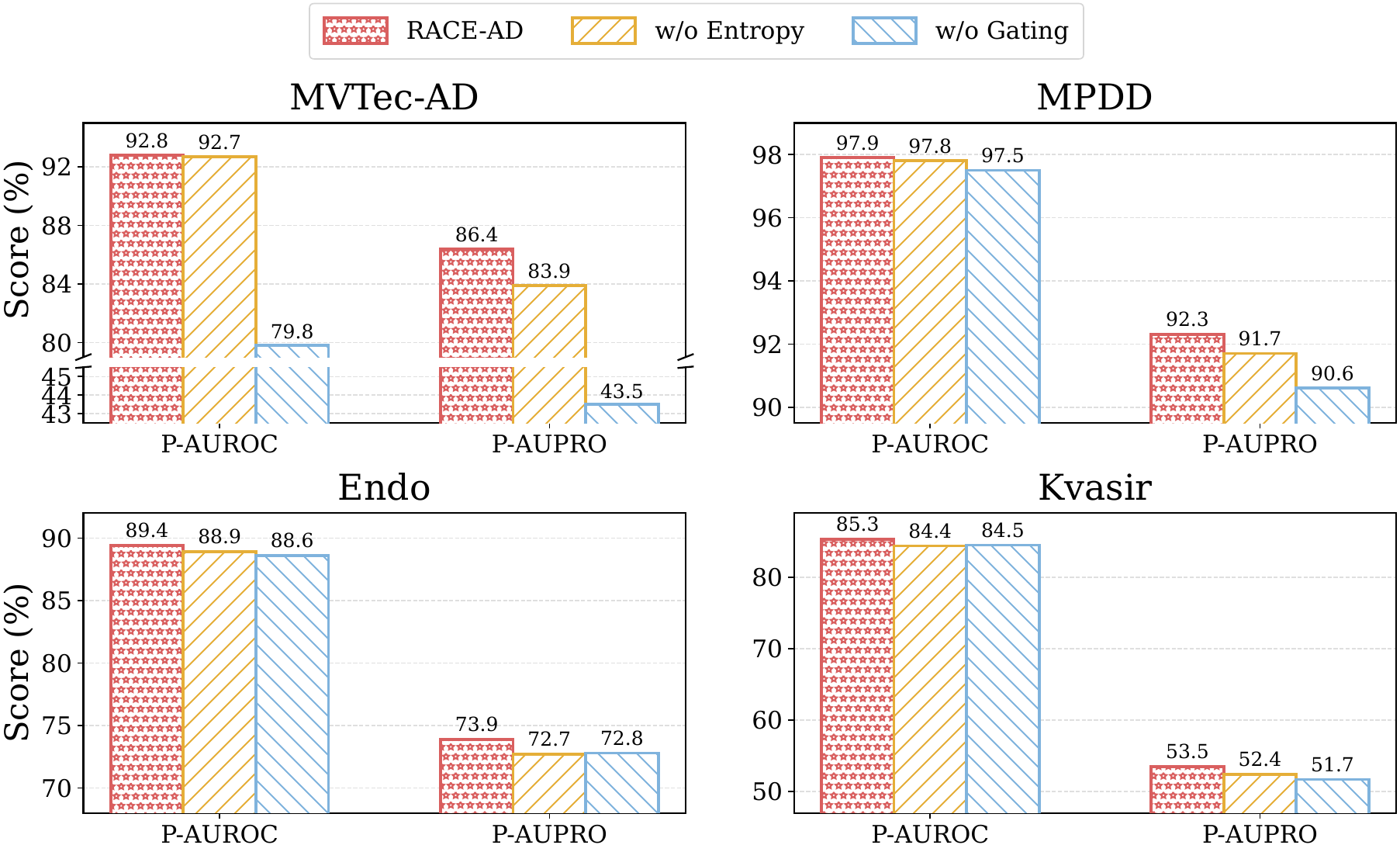}
    \caption{Ablation of entropy and gating on two industrial and two medical benchmarks, pixel-level AUROC and AUPRO (\%).}
    \label{fig:ablation}
\end{figure}

We study the two components on four representative benchmarks, MVTec-AD, MPDD, Endo, and Kvasir, which cover industrial and medical targets and both source-to-target transfers. Fig.~\ref{fig:ablation} reports pixel-level results of RACE-AD and of its variants without attention-derived entropy or without the confidence gate.

\begin{figure*}[t]
    \centering
    \begin{minipage}[t]{0.485\textwidth}
        \centering
        \includegraphics[width=\linewidth]{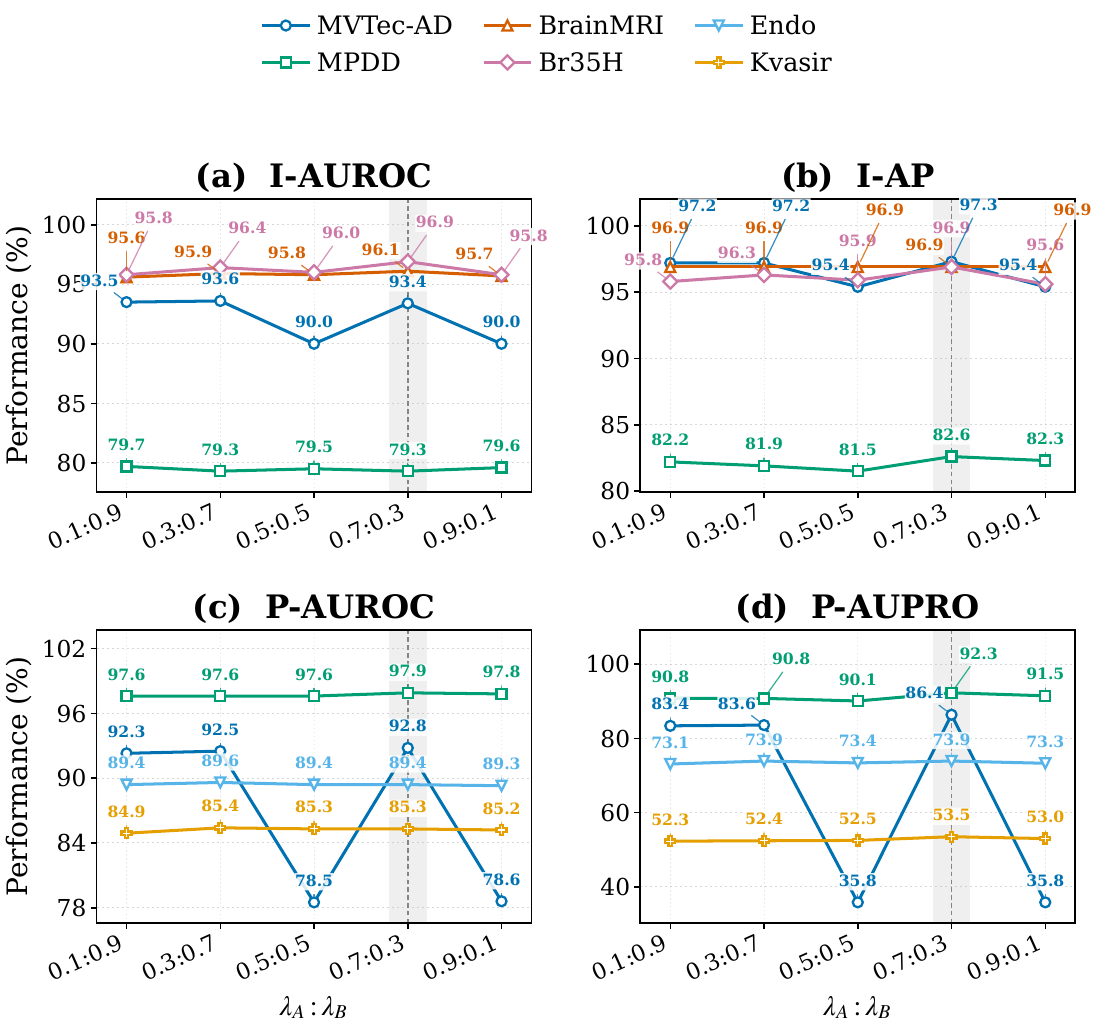}
        \caption{Sensitivity to the adapter loss weights $(\lambda_A,\lambda_B)$. The shaded band marks the setting used for all results.}
        \label{fig:sensitivity-lambda}
    \end{minipage}\hfill
    \begin{minipage}[t]{0.485\textwidth}
        \centering
        \includegraphics[width=\linewidth]{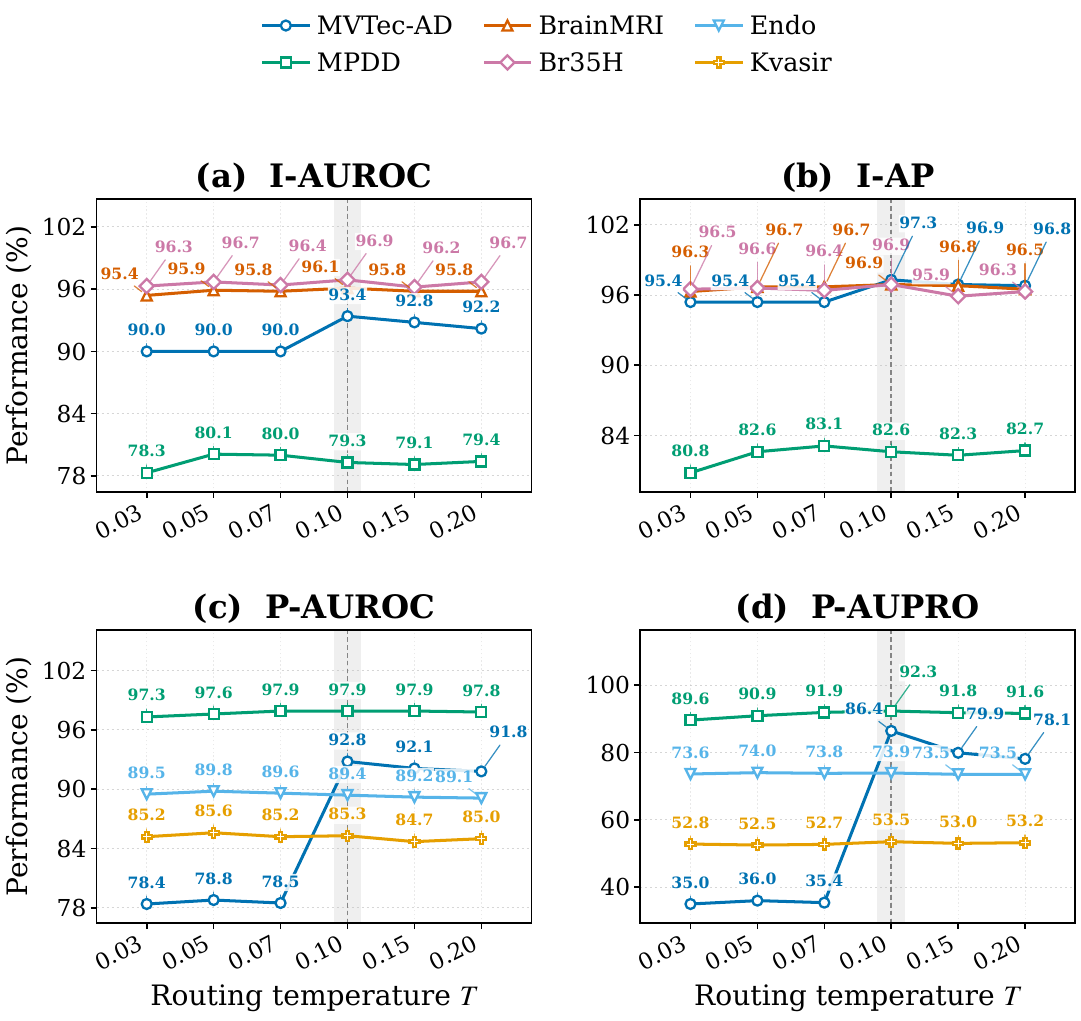}
        \caption{Sensitivity to the routing temperature $T$. The shaded band marks the setting used for all results.}
        \label{fig:sensitivity-T}
    \end{minipage}
\end{figure*}

\noindent\textbf{Effectiveness of Relation-Aware Evidence Routing.}
As shown in Fig.~\ref{fig:ablation}, removing attention-derived entropy from the routing reduces AUPRO on all four benchmarks, by 1.4 points on average and by 0.6 to 2.5 points per dataset, while pixel-level AUROC changes by at most 0.9 points. The entropy-guided routing therefore mainly improves region-level localization, which agrees with its role of selecting the patches that form the image-conditioned tokens.

\noindent\textbf{Effectiveness of Confidence-Guided Adaptation.}
Removing the confidence gate also reduces AUPRO on all four benchmarks, but the size of the effect depends on the transfer. On MPDD, Endo, and Kvasir, which are evaluated with the MVTec-AD-trained model, AUPRO decreases by 1.1 to 1.8 points. On MVTec-AD, evaluated with the VisA-trained model, AUPRO falls from 86.4\% to 43.5\% and pixel-level AUROC from 92.8\% to 79.8\%, which accounts for most of the 11.9-point average decrease. Without the gate, the routed anomaly token enters the adapters at full strength for every image, including images whose strongest memory response is weak. The collapse on MVTec-AD indicates that this transfer suffers most from unreliable anomaly evidence, which is the situation that the gate is designed to handle.

\begin{figure}[t]
    \centering
    \includegraphics[width=\linewidth]{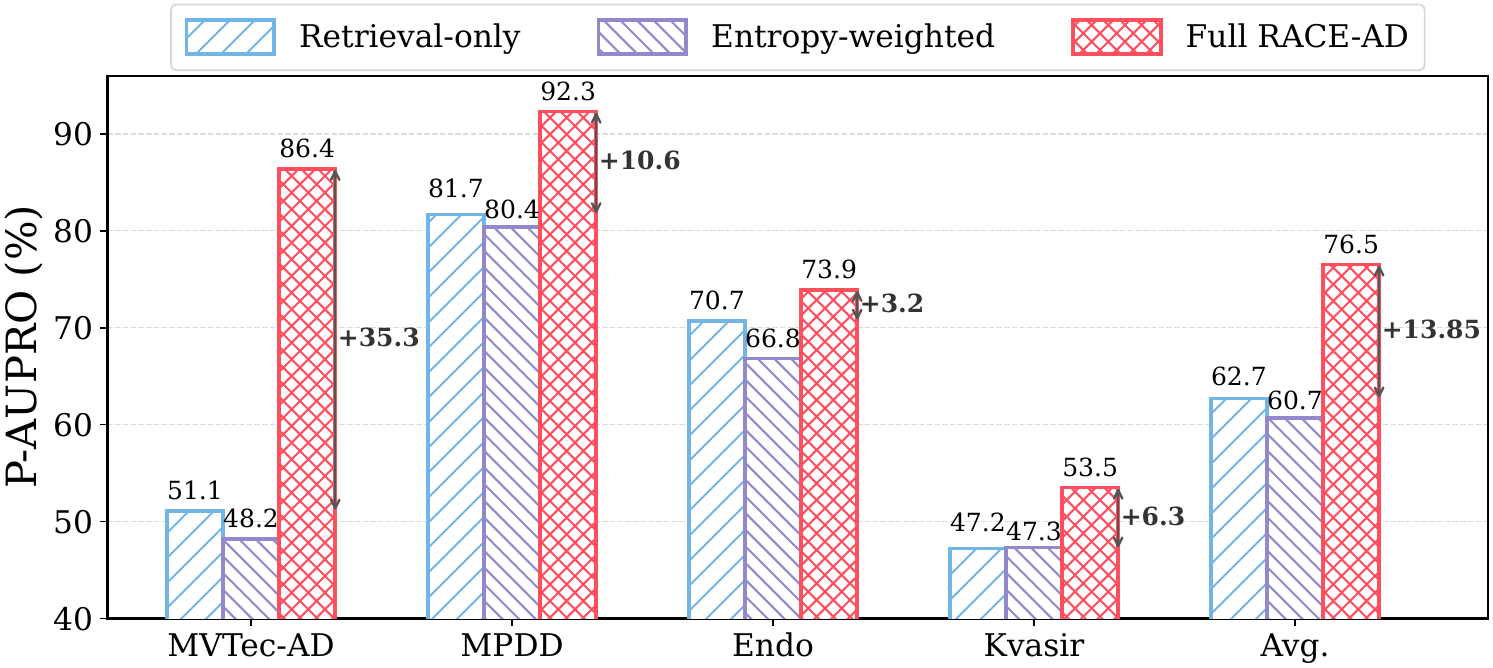}
    \caption{Pixel-level AUPRO (\%) of retrieval-only, entropy-weighted retrieval, and the full RACE-AD. Numbers in orange are the gains over retrieval-only.}
    \label{fig:retrieval-diagnostic}
\end{figure}

\subsection{Further Analysis}
\label{sec:analysis}

\noindent\textbf{Qualitative Visualization.}
Fig.~\ref{fig:qualitative_comparison} compares anomaly maps on hazelnut, cable, and toothbrush. RACE-AD follows the ground-truth regions more closely and responds less to the background, in line with its higher AUPRO.

\noindent\textbf{Hyperparameter Sensitivity.}
Figs.~\ref{fig:sensitivity-lambda} and~\ref{fig:sensitivity-T} vary the adapter loss weights $(\lambda_A,\lambda_B)$ and the routing temperature $T$ with all other settings fixed. For MPDD, BrainMRI, Br35H, Endo, and Kvasir, image-level metrics and AUPRO vary by less than three points across both sweeps, and no single configuration is best for all metrics. On MVTec-AD, AUPRO is 86.4\% at $T=0.1$, decreases to about 78--80\% for larger temperatures, and falls to about 35\% for $T\le0.07$; it also drops to about 36\% for two of the five loss-weight ratios. The drop at small temperatures is consistent with Proposition~\ref{prop:stability}, under which the sensitivity of the routed tokens to memory errors grows as $1/T$, and its size is comparable to the effect of removing the confidence gate on MVTec-AD (Fig.~\ref{fig:ablation}). Both observations identify the VisA-to-MVTec-AD transfer as the one most affected by unreliable anomaly evidence.

\noindent\textbf{Role of Relational Evidence.}
Fig.~\ref{fig:retrieval-diagnostic} compares three uses of the retrieval pathway: retrieval-only prediction, retrieval responses weighted by attention entropy, and the complete RACE-AD. Weighting the retrieval responses by entropy lowers the average AUPRO from 62.7\% to 60.7\%, so entropy is not a reliable anomaly score on its own. The complete framework reaches 76.5\%. Part of this gain comes from prompt adaptation, which the retrieval-only pathway does not use, whereas the ablation in \cref{sec:ablation} isolates the contribution of entropy within the complete framework. Together, the two analyses show that attention entropy is useful when it shapes the image-conditioned evidence and harmful when it rescales anomaly responses directly.

\noindent\textbf{Implementation Cost.}
RACE-AD trains 15.4M parameters in the image-level and patch-level memory projections, the prompt learner, and the two prompt adapters, while the CLIP encoders remain frozen. The attention entropy is computed from the self-attention matrices saved during the forward pass of the visual encoder, so each image passes through the visual encoder once, and the routing weights and the confidence gate are computed in closed form. Because the prompts are conditioned on the image, each of the two adapters runs the text encoder once per image. The keys and values of the source-domain memory occupy 16.7\,MiB when MVTec-AD is the source and 18.5\,MiB when VisA is the source. With the model and the memory loaded, inference with a batch size of 1 and $518\times518$ inputs reaches a peak PyTorch-allocated GPU memory of 4.52\,GiB.
\section{Conclusion}
\label{sec:conclusion}

We presented RACE-AD, a ZSAD framework that complements memory-based anomaly evidence with attention-derived relational statistics through relation-aware evidence routing and confidence-guided adaptation. Our analysis shows that the routing is optimal for an entropy-regularized selection objective, that attention entropy damps the effect of memory errors on the routed tokens, and that the gate bounds the influence of weak anomaly evidence on the prompts. RACE-AD obtains the best average detection and localization results on ten industrial and medical benchmarks.

\section*{LLM Usage Statement}
We used ChatGPT to assist with language refinement during the preparation of this manuscript. All AI-assisted content was reviewed and verified by the authors, who take full responsibility for the manuscript's content and experimental results.

{
    \small
    \bibliographystyle{ieeenat_fullname}
    \bibliography{main}
}

\maketitlesupplementary
\appendix
\raggedbottom

\makeatletter
\setlength{\@dblfptop}{0pt}
\setlength{\@dblfpsep}{10pt plus 2pt minus 2pt}
\setlength{\@dblfpbot}{0pt plus 1fil}
\makeatother

\setcounter{figure}{0}
\setcounter{table}{0}
\setcounter{equation}{0}
\setcounter{algocf}{0}
\setcounter{lemma}{0}
\renewcommand{\thetable}{S\arabic{table}}
\renewcommand{\thefigure}{S\arabic{figure}}
\renewcommand{\theequation}{S\arabic{equation}}
\renewcommand{\thealgocf}{S\arabic{algocf}}
\renewcommand{\thelemma}{S\arabic{lemma}}
\def\theHtable{S\arabic{table}}
\def\theHfigure{S\arabic{figure}}
\def\theHequation{S\arabic{equation}}
\def\theHalgocf{S\arabic{algocf}}
\def\theHlemma{S\arabic{lemma}}

\section{Proofs}
\label{sec:supp_proof}

Throughout, $\mathbf{p},\hat{\mathbf{e}}\in[0,1]^N$, $T>0$, and $\mathbf{w}^c$ is the spatial softmax of the utilities $\mathbf{q}^c$ in Eq.~\eqref{eq:routing_weights}. We use the convention $0\log0=0$.

\subsection{Evidence Agreement}

\begin{proof}[Proof of Proposition~\ref{prop:agreement}]
Since $p_i,\hat e_i\in[0,1]$, we have $q_i^a=p_i\hat e_i\le p_i$ and $q_i^a\le\hat e_i$, and likewise $q_i^n=(1-p_i)(1-\hat e_i)\le\min(1-p_i,1-\hat e_i)$. Expanding $q_i^n=1-p_i-\hat e_i+p_i\hat e_i$ gives $q_i^a-q_i^n=p_i+\hat e_i-1$.
\end{proof}

\subsection{Optimal Evidence Routing}

\begin{proof}[Proof of Proposition~\ref{prop:routing}]
Fix $c\in\{n,a\}$ and write $q_i=q_i^c$. The first term of $J$ is linear in $\mathbf{w}$, and the entropy $\mathcal{H}(\mathbf{w})=\sum_i\phi(w_i)$ with $\phi(t)=-t\log t$ is a sum of strictly concave functions, because $\phi''(t)=-1/t<0$ for $t>0$ and $\phi$ is continuous at $t=0$. Hence $J$ is strictly concave on the convex set $\Delta_N$ and has at most one maximizer; since $\Delta_N$ is compact and $J$ is continuous, a maximizer exists.

Let $Z=\sum_{j=1}^{N}\exp(q_j/T)$ and $w_i^{*}=\exp(q_i/T)/Z$. Then $w_i^{*}>0$ and $q_i=T\log w_i^{*}+T\log Z$ for every $i$. For any $\mathbf{w}\in\Delta_N$, using $\sum_iw_i=1$,
\begin{align}
J(\mathbf{w})
&=\sum_{i}w_i\bigl(T\log w_i^{*}+T\log Z\bigr)-T\sum_{i}w_i\log w_i \nonumber\\
&=T\log Z-T\sum_{i}w_i\log\frac{w_i}{w_i^{*}} \nonumber\\
&=T\log Z-T D_{\mathrm{KL}}(\mathbf{w}\,\|\,\mathbf{w}^{*}).
\label{eq:supp_gap}
\end{align}
Setting $\mathbf{w}=\mathbf{w}^{*}$ gives $J(\mathbf{w}^{*})=T\log Z$, so that $J(\mathbf{w}^{*})-J(\mathbf{w})=T D_{\mathrm{KL}}(\mathbf{w}\,\|\,\mathbf{w}^{*})\ge0$, with equality if and only if $\mathbf{w}=\mathbf{w}^{*}$ by Gibbs' inequality. The softmax is therefore the unique maximizer, $\mathbf{w}^{c*}=\mathbf{w}^{*}$, which proves Eq.~\eqref{eq:routing_optimality}.
\end{proof}

Eq.~\eqref{eq:supp_gap} also shows that the optimal value is $T\log\sum_j\exp(q_j/T)$. As $T\to0$, $\mathbf{w}^{*}$ concentrates on $\arg\max_iq_i$, and as $T\to\infty$, it tends to the uniform distribution $\mathbf{1}/N$.

\subsection{Stability of Routed Tokens}

\begin{lemma}[Softmax sensitivity]\leavevmode\label{lem:softmax}%
For any $\mathbf{q},\mathbf{q}'\in\mathbb{R}^N$ with $\mathbf{v}=\mathbf{q}'-\mathbf{q}$, $\mathbf{w}=\operatorname{softmax}(\mathbf{q}/T)$, and $\mathbf{w}'=\operatorname{softmax}(\mathbf{q}'/T)$,
\begin{align}
\|\mathbf{w}'-\mathbf{w}\|_1
&\le\frac{1}{2T}\Bigl(\max_iv_i-\min_iv_i\Bigr)\nonumber\\
&\le\frac{1}{T}\max_i|v_i|.
\label{eq:supp_softmax}
\end{align}
\end{lemma}

\begin{proof}
Let $\mathbf{w}(s)=\operatorname{softmax}((\mathbf{q}+s\mathbf{v})/T)$ for $s\in[0,1]$. Differentiating the softmax gives
\begin{equation}
\frac{dw_i(s)}{ds}=\frac{1}{T}\,w_i(s)\bigl(v_i-\bar v(s)\bigr),
\quad
\bar v(s)=\sum_jw_j(s)v_j,
\end{equation}
so $\|d\mathbf{w}(s)/ds\|_1=T^{-1}\,\mathbb{E}_{i\sim\mathbf{w}(s)}|v_i-\bar v(s)|$ is the mean absolute deviation of $v_i$ under $\mathbf{w}(s)$, divided by $T$. By Jensen's inequality and Popoviciu's inequality on variances, $\mathbb{E}|v_i-\bar v|\le\sqrt{\operatorname{Var}(v_i)}\le(\max_jv_j-\min_jv_j)/2$. Integrating over $s\in[0,1]$ gives the first inequality, and $\max_jv_j-\min_jv_j\le2\max_j|v_j|$ gives the second.
\end{proof}

\begin{proof}[Proof of Proposition~\ref{prop:stability}]
For the anomaly branch, $q_i'^a-q_i^a=\hat e_i\delta_i$, so Lemma~\ref{lem:softmax} gives $\|\mathbf{w}'^a-\mathbf{w}^a\|_1\le T^{-1}\max_i\hat e_i|\delta_i|$. Let $\mathbf{c}$ be the center of the smallest ball of radius $R$ that contains $\{\mathbf{z}_i\}$. Both weight vectors sum to one, so $\sum_i(w_i'^a-w_i^a)=0$ and
\begin{align}
\|\mathbf{t}_a'^{\,\mathrm{raw}}-\mathbf{t}_a^{\mathrm{raw}}\|_2
&=\Bigl\|\sum_i(w_i'^a-w_i^a)(\mathbf{z}_i-\mathbf{c})\Bigr\|_2\nonumber\\
&\le R\,\|\mathbf{w}'^a-\mathbf{w}^a\|_1.
\end{align}
For the normal branch, $q_i'^n-q_i^n=-(1-\hat e_i)\delta_i$, and the same argument applies.
\end{proof}

\subsection{Gate Suppression}

\begin{proof}[Proof of Proposition~\ref{prop:gate}]
Write $u=(p_{\max}-\tau)(k_0+k_1\sigma_e)$, so that $g=\sigma(u)$ with $\sigma'(u)>0$. Since $k_0,k_1>0$ and $\sigma_e\ge0$, we have $k_0+k_1\sigma_e>0$, hence $\partial g/\partial p_{\max}=\sigma'(u)(k_0+k_1\sigma_e)>0$, and $\partial g/\partial\sigma_e=\sigma'(u)\,k_1(p_{\max}-\tau)$ has the sign of $p_{\max}-\tau$. If $p_{\max}\le\tau-\Delta$, then $u\le-\Delta(k_0+k_1\sigma_e)$, and since $\sigma$ is increasing, $g\le\sigma(-\Delta(k_0+k_1\sigma_e))\le\sigma(-\Delta k_0)$. Finally, $\mathbf{t}_a^{\mathrm{raw}}$ is a convex combination of the features, so $\|\mathbf{t}_a^{\mathrm{raw}}\|_2\le B$, and since $\mathbf{b}_a^m-\mathbf{c}_a^m=g\,\mathbf{W}_a^m\mathbf{t}_a^{\mathrm{raw}}$,
\begin{equation}
\|\mathbf{b}_a^m-\mathbf{c}_a^m\|_2\le g\,\|\mathbf{W}_a^m\|_2\,B,
\end{equation}
which gives Eq.~\eqref{eq:gate_bound}.
\end{proof}

\section{Datasets and Evaluation Protocol}
\label{sec:supp_datasets}

\begin{table*}[!t]
  \centering
  \caption{The ten datasets used for evaluation. Target datasets are evaluated on their official test splits under the cross-dataset ZSAD protocol.}
  \label{tab:dataset_details}
  \begingroup
  \small
  \setlength{\tabcolsep}{5.8pt}
  \renewcommand{\arraystretch}{1.05}
  \begin{tabular*}{\textwidth}{@{\extracolsep{\fill}}llllcc@{}}
    \toprule
    \textbf{Domain}
    & \textbf{Dataset}
    & \textbf{Type}
    & \textbf{Modality}
    & \textbf{Categories}
    & \textbf{Test Samples (Normal / Anomaly)} \\
    \midrule

    \multirow{5}{*}{Industrial}
      & MVTec-AD~\cite{bergmann2019mvtec}
      & Object \& Texture
      & Photography
      & 15
      & 467 / 1258 \\

      & VisA~\cite{zou2022spot}
      & Object
      & Photography
      & 12
      & 962 / 1200 \\

      & MPDD~\cite{jezek2021deep}
      & Object
      & Photography
      & 6
      & 176 / 282 \\

      & BTAD~\cite{mishra2021vt}
      & Object
      & Photography
      & 3
      & 451 / 290 \\

      & DTD~\cite{aota2023zero}
      & Texture
      & Photography
      & 12
      & 357 / 947 \\

    \midrule

    \multirow{5}{*}{Medical}
      & HeadCT~\cite{salehi2021multiresolution}
      & Brain
      & Radiology (CT)
      & 1
      & 100 / 100 \\

      & BrainMRI~\cite{kanade2015brain}
      & Brain
      & Radiology (MRI)
      & 1
      & 98 / 155 \\

      & Br35H~\cite{hamada2020br35h}
      & Brain
      & Radiology (MRI)
      & 1
      & 1500 / 1500 \\

      & Kvasir~\cite{jha2019kvasir}
      & Colon
      & Endoscopy
      & 1
      & 0 / 1000 \\

      & Endo~\cite{hicks2021endotect}
      & Colon
      & Endoscopy
      & 1
      & 0 / 200 \\

    \bottomrule
  \end{tabular*}
  \endgroup
\end{table*}

We evaluate \modelname on ten public datasets covering industrial defect inspection and medical image analysis. Under the cross-dataset ZSAD protocol, the model is trained only on the source dataset and evaluated directly on the test sets of unseen target datasets, without any target-domain adaptation. Table~\ref{tab:dataset_details} lists the imaging modality, the number of categories, and the numbers of normal and anomalous test samples of each dataset. Endo and Kvasir contain only anomalous images and are therefore used for pixel-level evaluation only, while BrainMRI, HeadCT, and Br35H provide image-level labels only.

\section{Implementation Details}
\label{sec:supp_implementation}

\noindent\textbf{Architecture.}
Images and masks are resized to $518\times518$, which gives $N=37\times37$ patches for the CLIP ViT-L/14@336px encoder. Following AnomalyCLIP~\cite{zhouanomalyclip}, layers 2 to 24 of the visual encoder carry an additional value-value attention path that produces the dense patch features, while the global feature and the attention matrices used for the entropy come from the original path. Attention entropy uses the layers $\Omega=\{6,12,18,24\}$. Memory retrieval, routing, and prompt-based localization use the patch features of layer 24, averaged over $1\times1$, $3\times3$, and $5\times5$ neighborhoods and $\ell_2$-normalized. The prompt learner also follows AnomalyCLIP: the normal and anomaly prompts are ``$[V]_1\cdots[V]_{12}$ object.'' and ``$[V]_1\cdots[V]_{12}$ damaged object.'' with 12 learnable context embeddings each, and in layers 2 to 9 of the text encoder the first four tokens after the start token are replaced by learnable embeddings. Each prompt adapter consists of two linear layers that map $\mathbf{t}_n$ and $\mathbf{t}_a$ to 768-dimensional prompt biases, which are added to all 12 context embeddings of the corresponding prompt.

\noindent\textbf{Memory.}
The image-level memory stores the global features of the source images with one-hot image labels. In the patch-level memory, each source image contributes the mean feature of its anomalous patches and the mean feature of its normal patches as prototypes with one-hot values. During training, prototypes whose cosine similarity to the query exceeds the 80th percentile, or the 95th percentile for the image-level memory, are excluded; at inference, all prototypes are used.

\noindent\textbf{Training.}
Stage~1 is trained for one epoch and Stage~2 for five epochs with Adam ($\beta_1=0.5$, $\beta_2=0.999$), a learning rate of $4\times10^{-4}$, a batch size of 8, and the random seed 111. Stage~1 optimizes the query and key projections of the two memories, and Stage~2 optimizes the prompt learner and the adapters. The Dice weight is $\lambda_d=1$, the adapter loss weights are $\lambda_A=0.7$ and $\lambda_B=0.3$, and the routing temperature is $T=0.1$.

\noindent\textbf{Inference and Post-processing.}
The confidence gate in Eq.~\eqref{eq:confidence_gate} uses $\tau=0.5$, $k_0=5.0$, and $k_1=50.0$, where $\sigma_e$ is the standard deviation of $\hat{\mathbf{e}}$ over the patches of the image. The anomaly map of each adapter is smoothed with a Gaussian filter of standard deviation $\sigma=4$, and the two maps are averaged with equal weights. The image-level score in Eq.~\eqref{eq:final_score} combines the mean of the top-$1\%$ localization responses with the image-level retrieval score using $\kappa=0.7$.

\section{Training Procedure}
\label{sec:supp_procedure}

Algorithm~\ref{alg:training} summarizes the two-stage training of \modelname.

\begin{algorithm}[!t]
\caption{Two-stage training of \modelname}
\label{alg:training}
\KwIn{Source-domain training set $\mathcal{D}=\{(x,y,\mathbf{Y})\}$}
\KwOut{Trained \modelname}
\tcp{Stage 1: memory-guided retrieval}
Freeze the CLIP visual backbone\;
Optimize the projection and memory retrieval modules with $\mathcal{L}_{\mathrm{stage1}}$ in Eq.~\eqref{eq:stage1_loss}\;
\tcp{Stage 2: relation-aware routing and prompt adaptation}
Freeze the visual backbone and the Stage-1 modules\;
\For{each minibatch $(x,y,\mathbf{Y})\in\mathcal{D}$}{
  Extract the dense features $\{\mathbf{z}_i\}_{i=1}^{N}$, the projected features $\{\mathbf{r}_i\}_{i=1}^{N}$, and the attention matrices $\{\mathbf{A}^{(\ell)}\}_{\ell\in\Omega}$\;
  Compute the anomaly evidence $\{p_i\}$ with Eq.~\eqref{eq:memory_response}\;
  Compute the normalized attention entropy $\hat{\mathbf{e}}$ with Eqs.~\eqref{eq:attention}, \eqref{eq:attention_entropy}, and~\eqref{eq:entropy_normalization}\;
  Compute the routing weights and the tokens $(\mathbf{t}_n,\mathbf{t}_a^{\mathrm{raw}})$ with Eqs.~\eqref{eq:routing_weights} and~\eqref{eq:routed_tokens}\;
  Regulate the anomaly token, $\mathbf{t}_a=g\,\mathbf{t}_a^{\mathrm{raw}}$, with Eq.~\eqref{eq:confidence_gate}\;
  For $m\in\{A,B\}$, compute the prompt biases, the text representations, and the anomaly maps with Eqs.~\eqref{eq:adapted_prompts} and~\eqref{eq:patch_text_alignment}\;
  Update the prompt learner and the two adapters with $\mathcal{L}_{\mathrm{stage2}}$ in Eqs.~\eqref{eq:branch_loss} and~\eqref{eq:stage2_loss}\;
}
\Return{trained model}\;
\end{algorithm}

\section{Per-Category Results}
\label{sec:supp_results}

Tables~\ref{tab:detailed_mvtec_visa} and~\ref{tab:detailed_dtd_mpdd_btad} report per-category results on the five multi-class industrial benchmarks: image-level AUROC (I-AUROC) and average precision (I-AP), and pixel-level AUROC (P-AUROC) and AUPRO (P-AUPRO). The medical datasets contain a single category each, and their results are given in Tables~\ref{tab:zero-shot-image} and~\ref{tab:zero-shot-pixel} of the main paper.

\begin{table*}[!t]
  \centering
  \caption{Per-category results on MVTec-AD (evaluated with the VisA-trained model) and VisA (\%).}
  \label{tab:detailed_mvtec_visa}

  \begingroup
  \footnotesize
  \setlength{\tabcolsep}{2.5pt}
  \renewcommand{\arraystretch}{0.98}

  \begin{minipage}[t]{0.485\textwidth}
    \centering
    \textbf{(a) MVTec-AD}\par\vspace{2pt}

    \begin{tabular*}{\linewidth}{@{\extracolsep{\fill}}lcccc@{}}
      \toprule
      \multirow{2}{*}{\textbf{Category}}
      & \multicolumn{2}{c}{\textbf{Image-level}}
      & \multicolumn{2}{c}{\textbf{Pixel-level}} \\
      \cmidrule(lr){2-3}
      \cmidrule(lr){4-5}
      & I-AUROC & I-AP & P-AUROC & P-AUPRO \\
      \midrule

      \multicolumn{5}{l}{\textit{Textures}} \\
      Carpet     & 100.0 & 100.0 & 99.3 & 97.7 \\
      Grid       & 99.1  & 99.7  & 98.4 & 93.7 \\
      Leather    & 100.0 & 100.0 & 99.3 & 98.3 \\
      Tile       & 99.6  & 99.8  & 97.3 & 92.9 \\
      Wood       & 99.0  & 99.7  & 97.6 & 93.5 \\

      \midrule
      \multicolumn{5}{l}{\textit{Objects}} \\
      Bottle     & 94.1 & 98.3 & 91.1 & 83.7 \\
      Cable      & 95.8 & 97.7 & 79.5 & 68.4 \\
      Capsule    & 94.5 & 98.9 & 96.5 & 94.1 \\
      Hazelnut   & 96.9 & 98.5 & 96.3 & 93.1 \\
      Metal Nut  & 64.7 & 91.1 & 83.8 & 66.2 \\
      Pill       & 87.2 & 97.3 & 91.0 & 92.9 \\
      Screw      & 84.3 & 93.8 & 97.4 & 89.3 \\
      Toothbrush & 98.1 & 99.3 & 95.0 & 88.9 \\
      Transistor & 87.6 & 84.9 & 73.2 & 55.6 \\
      Zipper     & 99.8 & 99.9 & 96.7 & 88.5 \\

      \midrule
      \textbf{Average}
      & \textbf{93.4}
      & \textbf{97.3}
      & \textbf{92.8}
      & \textbf{86.4} \\
      \bottomrule
    \end{tabular*}
  \end{minipage}
  \hfill
  \begin{minipage}[t]{0.485\textwidth}
    \centering
    \textbf{(b) VisA}\par\vspace{2pt}

    \begin{tabular*}{\linewidth}{@{\extracolsep{\fill}}lcccc@{}}
      \toprule
      \multirow{2}{*}{\textbf{Category}}
      & \multicolumn{2}{c}{\textbf{Image-level}}
      & \multicolumn{2}{c}{\textbf{Pixel-level}} \\
      \cmidrule(lr){2-3}
      \cmidrule(lr){4-5}
      & I-AUROC & I-AP & P-AUROC & P-AUPRO \\
      \midrule

      Candle      & 84.1 & 87.7 & 99.0 & 96.3 \\
      Capsules    & 92.9 & 96.2 & 95.6 & 79.0 \\
      Cashew      & 93.1 & 97.1 & 94.3 & 94.7 \\
      Chewinggum  & 98.6 & 99.4 & 99.6 & 94.0 \\
      Fryum       & 93.0 & 96.9 & 92.7 & 91.1 \\
      Macaroni1   & 80.3 & 82.0 & 98.6 & 92.6 \\
      Macaroni2   & 73.7 & 73.0 & 97.0 & 81.6 \\
      PCB1        & 75.9 & 79.1 & 90.5 & 81.8 \\
      PCB2        & 67.1 & 70.1 & 90.7 & 74.0 \\
      PCB3        & 68.6 & 72.5 & 88.2 & 79.6 \\
      PCB4        & 96.9 & 96.9 & 95.7 & 87.1 \\
      Pipe Fryum  & 92.3 & 96.2 & 96.3 & 95.0 \\

      \midrule
      \textbf{Average}
      & \textbf{84.7}
      & \textbf{87.2}
      & \textbf{94.8}
      & \textbf{87.2} \\
      \bottomrule
    \end{tabular*}
  \end{minipage}

  \endgroup
\end{table*}

\begin{table*}[!t]
  \centering
  \caption{Per-category results on DTD, MPDD, and BTAD (\%).}
  \label{tab:detailed_dtd_mpdd_btad}

  \begingroup
  \footnotesize
  \setlength{\tabcolsep}{2.5pt}
  \renewcommand{\arraystretch}{0.98}

  \begin{minipage}[t]{0.485\textwidth}
    \centering
    \textbf{(a) DTD}\par\vspace{2pt}

    \begin{tabular*}{\linewidth}{@{\extracolsep{\fill}}lcccc@{}}
      \toprule
      \multirow{2}{*}{\textbf{Category}}
      & \multicolumn{2}{c}{\textbf{Image-level}}
      & \multicolumn{2}{c}{\textbf{Pixel-level}} \\
      \cmidrule(lr){2-3}
      \cmidrule(lr){4-5}
      & I-AUROC & I-AP & P-AUROC & P-AUPRO \\
      \midrule

      Woven 001      & 100.0 & 100.0 & 99.8 & 96.1 \\
      Woven 127      & 95.4  & 96.7  & 96.2 & 92.5 \\
      Woven 104      & 98.9  & 99.7  & 97.1 & 92.4 \\
      Stratified 154 & 98.4  & 99.6  & 99.7 & 94.6 \\
      Blotchy 099    & 98.5  & 99.6  & 99.6 & 96.9 \\
      Woven 068      & 94.3  & 96.9  & 98.9 & 92.3 \\
      Woven 125      & 100.0 & 100.0 & 99.5 & 97.5 \\
      Marbled 078    & 99.1  & 99.8  & 99.3 & 97.8 \\
      Perforated 037 & 91.2  & 97.9  & 95.5 & 79.9 \\
      Mesh 114       & 85.8  & 94.5  & 97.2 & 85.6 \\
      Fibrous 183    & 98.8  & 99.7  & 99.5 & 98.5 \\
      Matted 069     & 90.1  & 97.4  & 99.3 & 86.0 \\

      \midrule
      \textbf{Average}
      & \textbf{95.9}
      & \textbf{98.5}
      & \textbf{98.5}
      & \textbf{92.5} \\
      \bottomrule
    \end{tabular*}
  \end{minipage}
  \hfill
  \begin{minipage}[t]{0.485\textwidth}
    \centering
    \textbf{(b) MPDD}\par\vspace{2pt}

    \begin{tabular*}{\linewidth}{@{\extracolsep{\fill}}lcccc@{}}
      \toprule
      \multirow{2}{*}{\textbf{Category}}
      & \multicolumn{2}{c}{\textbf{Image-level}}
      & \multicolumn{2}{c}{\textbf{Pixel-level}} \\
      \cmidrule(lr){2-3}
      \cmidrule(lr){4-5}
      & I-AUROC & I-AP & P-AUROC & P-AUPRO \\
      \midrule

      Bracket Black & 68.0 & 79.3 & 98.4 & 95.4 \\
      Bracket Brown & 60.7 & 77.3 & 96.3 & 82.0 \\
      Bracket White & 76.1 & 72.1 & 99.5 & 98.1 \\
      Connector     & 77.9 & 69.6 & 97.6 & 90.9 \\
      Metal Plate   & 95.9 & 98.5 & 97.5 & 93.8 \\
      Tubes         & 97.0 & 98.8 & 98.3 & 93.7 \\

      \midrule
      \textbf{Average}
      & \textbf{79.3}
      & \textbf{82.6}
      & \textbf{97.9}
      & \textbf{92.3} \\
      \bottomrule
    \end{tabular*}

    \vspace{4mm}

    \textbf{(c) BTAD}\par\vspace{2pt}

    \begin{tabular*}{\linewidth}{@{\extracolsep{\fill}}lcccc@{}}
      \toprule
      \multirow{2}{*}{\textbf{Category}}
      & \multicolumn{2}{c}{\textbf{Image-level}}
      & \multicolumn{2}{c}{\textbf{Pixel-level}} \\
      \cmidrule(lr){2-3}
      \cmidrule(lr){4-5}
      & I-AUROC & I-AP & P-AUROC & P-AUPRO \\
      \midrule

      01 & 95.3 & 98.2 & 96.2 & 68.0 \\
      02 & 85.9 & 97.7 & 95.4 & 62.4 \\
      03 & 97.1 & 86.1 & 96.9 & 92.6 \\

      \midrule
      \textbf{Average}
      & \textbf{92.8}
      & \textbf{94.0}
      & \textbf{96.2}
      & \textbf{74.3} \\
      \bottomrule
    \end{tabular*}
  \end{minipage}

  \endgroup
\end{table*}

\end{document}